\documentclass[final]{cvpr}

\usepackage{times}
\usepackage{epsfig}
\usepackage{graphicx}
\usepackage{amsmath}
\usepackage{amssymb}

\usepackage{amsmath}
\usepackage{amssymb}
\usepackage{algorithm}
\usepackage{algorithmic}
\usepackage{subfigure}

\usepackage[pagebackref=true,breaklinks=true,colorlinks,bookmarks=false]{hyperref}

\def\cvprPaperID{4620} 
\def\confYear{CVPR 2021}

\begin{document}

\title{Cross-Domain Gradient Discrepancy Minimization \\for Unsupervised Domain Adaptation}
\author{Zhekai Du$^1$, Jingjing Li \thanks{Jingjing Li is the corresponding author.} $^1$, Hongzu Su$^1$, Lei Zhu$^2$, Ke Lu$^1$\\
      $^1$University of Electronic Science and Technology of China; $^2$Shandong Normal Unversity\\
      {\tt\small \{zhekaid, hongzus\}@std.uestc.edu.cn, lijin117@yeah.net, leizhu0608@gmail.com, kel@uestc.edu.cn}
   }
\maketitle
\pagestyle{empty}
\thispagestyle{empty}
\begin{abstract}
   Unsupervised Domain Adaptation (UDA) aims to generalize the knowledge learned from a well-labeled source domain to an unlabeled target domain. Recently, adversarial domain adaptation with two distinct classifiers (bi-classifier) has been introduced into UDA which is effective to align distributions between different domains. Previous bi-classifier adversarial learning methods only focus on the similarity between the outputs of two distinct classifiers. However, the similarity of the outputs cannot guarantee the accuracy of target samples, i.e., target samples may match to wrong categories even if the discrepancy between two classifiers is small. To challenge this issue, in this paper, we propose a cross-domain gradient discrepancy minimization (CGDM) method which explicitly minimizes the discrepancy of gradients generated by source samples and target samples. Specifically, the gradient gives a cue for the semantic information of target samples so it can be used as a good supervision to improve the accuracy of target samples. In order to compute the gradient signal of target samples, we further obtain target pseudo labels through a clustering-based self-supervised learning. Extensive experiments on three widely used UDA datasets show that our method surpasses many previous state-of-the-arts.
   \end{abstract}
   
   \vspace{-10pt}
   \section{Introduction}
   Conventional deep learning methods suffer from the challenge of heavy dependency on large-scale labeled data, which is extremely expensive in many real-world scenarios such as medical image analysis. To avoid expensive data annotation, unsupervised domain adaptation (UDA) \cite{pan2010domain, li2018heterogeneous, li2021faster} attempts to transfer a model trained on labeled data collected in the source domain to a similar target domain with unlabeled data. To mitigate the domain shift, one popular paradigm in UDA is to reduce the distribution divergence between domains by minimizing a specific metric \cite{long2015learning,long2017deep,zellinger2017central}. Another widely used paradigm aims to learn domain-invariant feature representations by leveraging the idea of adversarial learning \cite{ganin2015unsupervised}, which has achieved remarkable success in the field of UDA recently.
   
   Existing adversarial domain adaptation methods can be implemented in two ways. One way is to apply an extra domain discriminator to distinguish whether a sample comes from the source or the target domain. At the same time, a feature extractor is used to fool the domain discriminator by learning undistinguishable features from input samples \cite{ganin2016domain,ganin2015unsupervised,long2018conditional}. However, these domain adversarial methods neglect the category information of target samples, which may result in deterioration of the feature discriminability \cite{chen2019transferability}. Another adversarial paradigm proposes a within-network adversarial strategy with two classifiers \cite{saito2018maximum,lee2019sliced}. Through the minimax game between the classifiers and the generator on the cross-classifier outputs discrepancy, the target samples outside the support of the source domain can be detected by the decision boundaries effectively, thus the feature alignment could be established while the discriminability is also preserved. 

   \begin{figure}[t]
    \begin{center}
    \includegraphics[width=0.8\linewidth]{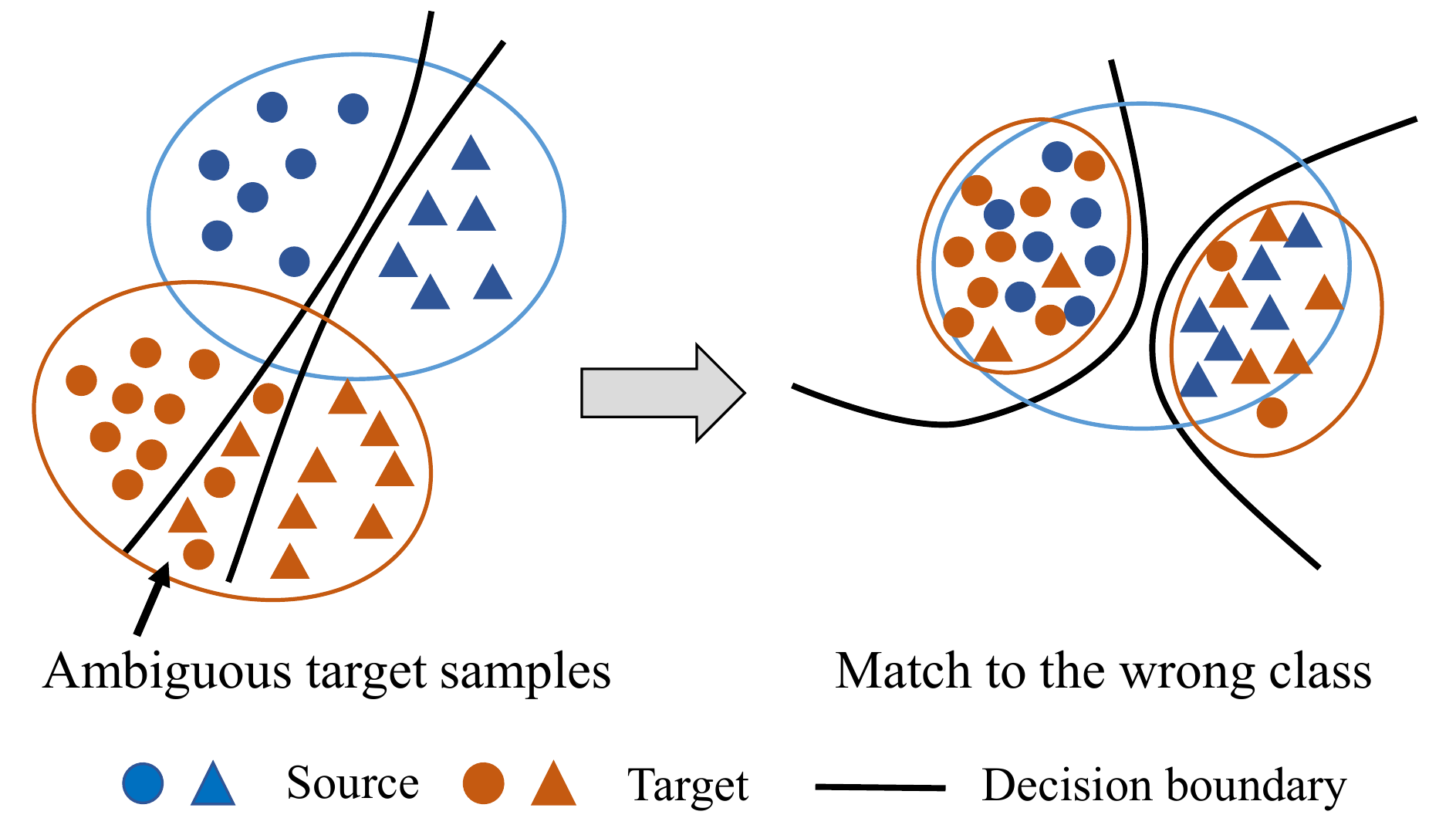}
    \end{center}
    \vspace{-13pt}
    \caption{Illustration of the issue in previous bi-classifier adversarial learning. Previous methods only consider the discrepancy between classifiers and neglect the accuracy of the target samples.}
    \label{fig:previous2}
    \vspace{-16pt}
   \end{figure}

   Although bi-classifier adversarial learning has shown promising performance, methods with this paradigm only focus on the similarity between two distinct classifiers through a discrepancy metric such as $\ell_1$ distance \cite{saito2018maximum} and slide wasserstein distance \cite{lee2019sliced}. Here we argue that only considering the discrepancy between classifiers cannot guarantee the accuracy and diversity of classification on target samples, because it is possible that both classifiers get wrong results. As shown in Fig. \ref{fig:previous2}, ambiguous target samples may detected by wrong decision boundaries, which inevitably results in an inaccurate class-wise distribution alignment in subsequent adversarial procedure, although the discrepancy metric, e.g., $\ell_1$ and wasserstein distance between the outputs of two classifiers is small. The main reason is that previous bi-classifier adversarial methods lack the consideration of the accuracy on target samples.
   
   In this paper, we aim to alleviate this issue. One straightforward idea is to label target samples by pseudo labels and fine-tune the model with data of both domains, which has been proven to be effctive in UDA \cite{choi2019pseudo}. However, directly using hard pseudo labels for supervised learning would lead to error diffusion and converge to the accuracy of pseudo labels. Here we tackle this issue from another perspective, we notice that the gradient discrepancy between source and target samples is related to the accuracy: assuming there is an accurate classifier, the source data and the target data would produce similar gradient signals for updating the classifier. Our key idea is that we wish the loss of two domains to be close to not only the final model but also to follow a similar path to it throughout the optimization. Therefore, we use the gradient signal as a surrogate supervision and make it an proxy to the classification accuracy, yielding a cross-domain gradient discrepancy minimization (CGDM) method for UDA. CGDM employs the gradient discrepancy between source and target samples as an extra supervision. Furthermore, considering that pseudo labels obtained by the source-only classifier may be not accurate enough, we leverage a clustering-based self-supervised method to obtain more reliable pseudo labels for target samples. By aligning gradient vectors, distributions of two domains can be better aligned at the category-level. The main contributions of this work are summarized as follows:
   \begin{itemize} 
       \vspace{-8pt}
       \item In order to solve the inaccurate alignment issue in previous bi-classifier adversarial learning, we propose a novel method which explicitly minimizes the discrepancy of gradient vectors produced by the source and the target samples. Notably, we formulate the proposed gradient discrepancy minimization as a generalized learning loss which can be easily applied to other UDA paradigms.
       \vspace{-8pt}
       \item For computing the gradient of target samples, we employ a clustering-based strategy to obtain more reliable pseudo labels. Then a self-supervised learning based on pseudo labels is conducted to fine-tune the model with both the source data and the target data in order to reduce the number of ambiguous target samples. 
       \vspace{-8pt}
       \item We reformulate the vanilla bi-classifier adversarial framework with above two proposals and conduct extensive experiments on three open large-scale datasets. The experimental results demonstrate the advantage of our method. 
   \end{itemize}
   
   \begin{figure*}[t]
       \begin{center}
       \includegraphics[width=0.77\linewidth]{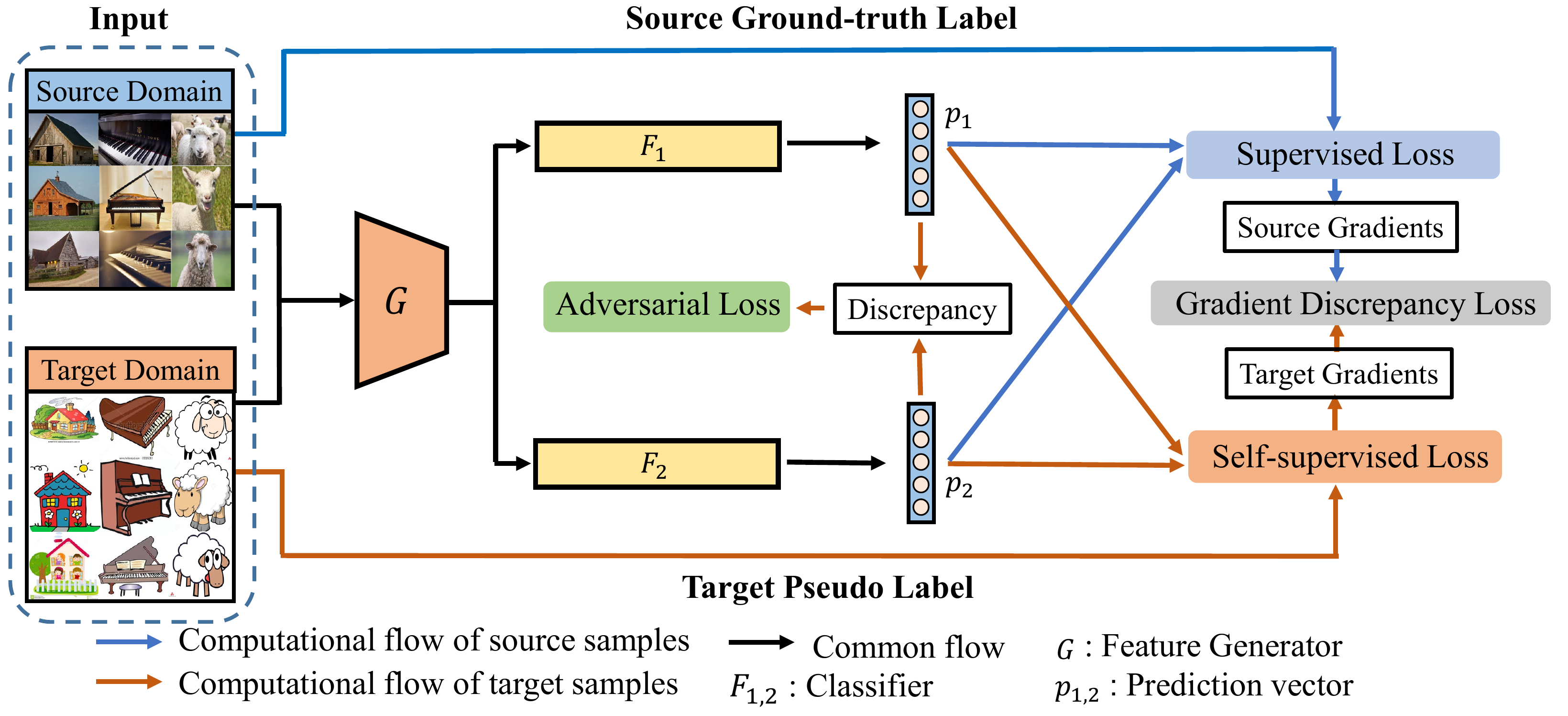}
       \end{center}
       \vspace{-10pt}
       \caption{An illustration of our framework. Both source samples and target samples are passed through the generator $G$ and two classifiers $F_1$ and $F_2$. Then the supervised loss is used to minimize the classification error on source samples. We use a self-supervised mechanism to reduce the number of ambiguous target samples. The minimax process on the adversarial loss is used to detect target samples outside the support of the source domain. The gradient discrepancy loss is further used to align distributions and improve the accuracy of target samples.}
       \label{fig:illustration}
       \vspace{-13pt}
   \end{figure*}
   \vspace{-10pt}
 
   \section{Related Work}
    
     The research line of existing unsupervised domain adaptation can be roughly divided into three branches: sample weighted adaptation, metric learning adaptation and adversarial learning adaptation. \textbf{Sample weighted adaptation} aims to reduce the discrepancy between domains by inferring resampling weights of samples in a non-parameter way \cite{wang2017instance,yan2019weighted, li2019locality, li2018transfer}. \textbf{Metric learning} methods try to mitigate the distribution gap directly by minimizing a discrepancy metric. For instance, maximum mean discrepancy (MMD) is a widely used criteria to measure the divergence between different domains in previous work. Deep adaptation networks (DAN) \cite{long2015learning} simultaneously minimize the multi-kernel MMD between two domains and the accuracy error on source samples, by which the distributions can be aligned. Joint adaptation networks (JAN) \cite{long2017deep} extends DAN by aligning the joint distributions of multiple domain-specific layers using the joint maximum mean discrepancy (JMMD). Besides, some other variants of MMD are also used for more appropriate divergence criteria \cite{long2016unsupervised,li2020deep, kang2019contrastive}. In addition, central moment discrepancy (CMD) \cite{zellinger2017central} and maximum density divergence (MDD) \cite{li2020maximum} are another two criterias to align feature distributions in hidden layers.  
     
   \textbf{Adversarial learning methods} learns domain-invariant feature representations following an adversarial paradigm. The adversarial learning of this paradigm can be realized through two strategies. The first way is to employ an additional domain discriminator to distinguish domain-specific features and a feature learner is then employed to learn undistinguishable features to fool the discriminator \cite{ganin2015unsupervised, tzeng2017adversarial}. Later, some studies \cite{long2018conditional, pei2018multi, zhang2019domain, li2019cycle} suggest to align conditional distributions in feature space to achieve accurate alignment at the category-level. Different from them, our method falls into the second adversarial paradigm, which uses two distinct task-specific classifiers to oppose the generator \textit{w.r.t.} the prediction discrepancy of two classifiers. This paradigm is first used in MCD \cite{saito2018maximum}. SWD \cite{lee2019sliced} further improves the discrepancy metric in MCD by applying the slide wasserstein discrepancy rather than simple $\ell_1$ distance. Recently, CLAN \cite{luo2019taking} use cosine similarity of the classifier parameters to measure the discrepancy for semantic segmentation. However, accurate alignment in this paradigm cannot be accessed by previous methods. 
 
   \textbf{Pseudo label based methods.} Some recent UDA methods use pseudo labeling technique to exploit semantic information of target samples. Zhang \textit{et al.} in \cite{zhang2018collaborative} directly use pseudo labels as a regularization. Xie \textit{et al.} \cite{xie2018learning} utilize pseudo labels to estimate class centroids for the target domain and match them to the ones in the source domain. Long \textit{et al.} \cite{long2018conditional} use pseudo labels predicted by the model to achieve conditional distribution alignment. Zou \textit{et al.} \cite{zou2018unsupervised} further propose a self-training framework that alternately refines pseudo labels and performs model training.  Recently, clustering-based pseudo labeling methods \cite{caron2018deep, liang2020we, kang2019contrastive} have shown their superiority and successfully applied to domain adaptation. CAN \cite{kang2019contrastive} solves the target pseudo labels by performing k-means clustering in feature space for contrastive learning. SHOT \cite{liang2020we} leverages a weighted k-means clustering strategy to obtain pseudo labels for self-supervision in the source-free UDA. Our method follow this strategy, instead of directly using pseudo labels for self-training, we propose to minimize the cross-domain gradient discrepancy based on pseudo labels to reinforce the vanilla bi-classifier adversarial learning, which is illustrated in Fig. \ref{fig:illustration} and elaborated on in subsequent sections.

   \section{Method}
   In this section, we first revisit the UDA and the bi-classifier adversarial setting. Then we introduce the cross-domain gradient discrepancy minimization and the self-supervised mechanism respectively. Finally, we report the whole training schema of our CGDM.
   
   \subsection{UDA with Bi-Classifier Adversarial Learning}
   
   Suppose that we have the source data which consists of $n_s$ labeled samples $\{\boldsymbol{x}_{i}^{s}\}_{i=1}^{n_s}$ and corresponding labels $\{{y}_{i}^{s}\}_{i=1}^{n_s}$ drawn from the source distribution $P(\mathcal{X}^{s},\mathcal{Y}^{s})$, as well as the target data which consists of $n_t$ unlabeled samples $\{\boldsymbol{x}_{i}^{t}\}_{i=1}^{n_t}$ drawn from the target distribution $P(\mathcal{X}^{t})$. UDA aims to obtain a function $\mathcal{F}$ that can predict the category of samples accurately with only source labels accessible and can be applied well to the target domain. 
   
   The bi-classifier adversarial learning methods employ a feature generator $G$ which extracts discriminative deep features of raw inputs, and two distinct task-specific classifiers $F_1$ and $F_2$, which are fed with the output of the generator then produce the prediction probability $p_{1}(\boldsymbol{y} \mid \boldsymbol{x}), p_{2}(\boldsymbol{y} \mid \boldsymbol{x})$ respectively. Following MCD \cite{saito2018maximum} and SWD \cite{lee2019sliced}, we first revisit the standard three-step principle of bi-classifier adversarial learning, which is used as our starting point: 
   
   \textbf{Step 1.} Learn $G, F_1$ and $F_2$ jointly by minimizing the classification loss $\mathcal{L}_{cls}(\cdot, \cdot)$ on the labeled source samples to reduce the empirical risk over the source distribution, which can be formulated as follows:
   \begin{equation}
       \begin{aligned}
   \min _{\theta_g, \theta_{f1}, \theta_{f2}} \mathcal{L}_{cls}\left(\mathcal{X}^{s}, \mathcal{Y}^{s}\right) = \frac{1}{2 n_{s}} \sum_{i=1}^{n_{s}} \sum_{n=1}^{2} \mathcal{L}_{ce}\left(F_{n}\left(G\left(\boldsymbol{x}_{i}^{s}\right)\right), y_{i}^{s}\right)
       \end{aligned}
   \end{equation}
   where $\mathcal{L}_{ce}(\cdot,\cdot)$ denotes the standard cross entropy loss function, $\theta_g, \theta_{f1}$ and $\theta_{f2}$ represent the parameters of $G$, $F_1$ and $F_2$ respectively. 
   
   \textbf{Step 2.} Frozen the parameters of the generator $G$ and update classifiers $F_1$ and $F_2$ to maximize the divergence between their probabilistic outputs on target samples while preserving the classification accuracy on source samples:
   
   \begin{equation}
       \begin{aligned}
   \min _{\theta_{f1}, \theta_{f2}} \mathcal{L}_{cls}\left(\mathcal{X}^{s}, \mathcal{Y}^{s}\right) - \mathcal{L}_{dis}\left(\mathcal{X}^{t}\right)
       \end{aligned}
   \end{equation}
   where $\mathcal{L}_{dis}(\cdot)$ denotes the function that measures the divergence between the probabilistic outputs of two classifiers and can be customized by specific algorithms like $\ell_1$ distance \cite{saito2017adversarial} and slide wasserstein distance \cite{lee2019sliced}.
   
   \textbf{Step 3.} Frozen the parameters of two classifiers $F_1$ and $F_2$, then update the generator $G$ to minimize the divergence between the probabilistic outputs of two classifiers:
   \begin{equation}
       \begin{aligned}
   \min _{\theta_{g}} \mathcal{L}_{dis}\left(\mathcal{X}^{t}\right)
       \end{aligned}
   \end{equation}
   After repeating above steps several times, the model can effectively detect target samples outside the support of the source domain by decision boundaries and align distributions of two domains to some extent.
   
   \subsection{Minimizing Cross-Domain Gradient Discrepancy}
   As mentioned above, it is difficult to guarantee that the model can classify the target sample with high accuracy if only considering the discrepancy of two classifiers without any other constraints. For instance, we have two predictions $[0.95,0.03,0.02]$ and $[0.96,0.02,0.02]$ of a target sample, one for each classifier. Although the discrepancy metric of two classifiers is quite small, we can not guarantee it is a good prediction because the vanilla bi-classifier adversarial learning is accuracy-agnostic, the groudtruth label of this sample may be [0,1,0] or [0,0,1]. Thus there is a strong incentive to tackle this issue.
   
   Based on the above motivation, we look at the problem from another perspective: if we want to learn a classifier through which all samples from both domains can be classified correctly, then the gradient vector produced by the samples from the source and the target should be similar for learning such a classifier. Therefore, we introduce the gradient similarity metric between two domains. To this end, we first denote the expected gradient over the source and target examples by $g_s$ and $g_t$ respectively. The appropriate gradient of source samples is formulated as follows,
   
   \begin{equation}
   g_{\mathrm{s}}= \frac{1}{2} \sum_{n=1}^{2} \underset{(\boldsymbol{x}_{i}^{s}, {y}_{i}^{s}) \sim\\ {(\mathcal{X}^{s},\mathcal{Y}^{s})}}{\mathbb{E}}\left[\nabla_{\theta_{fn}} \mathcal{L}_{\mathrm{ce}}\left(F_{n}\left(G\left(\boldsymbol{x}_{i}^{s}\right)\right), {y}_{i}^{s}\right)\right].
   \end{equation}
   For computing the gradient produced by target samples, the label information is required, however, this is exactly what we want to obtain in the UDA problem. To tackle this awkward situation, we assign pseudo labels to target samples, denoted by $\mathcal{Y}^{*}$. Since the pseudo labels may still incorrect, to alleviate the noise of ambiguous target samples which may have incorrect pseudo labels, we use the weighted classification loss based on the prediction entropy of each target sample to compute the gradient. The gradient vector of the target samples is formulated as follows,
   
   \begin{equation}
       g_{\mathrm{t}}= \frac{1}{2} \sum_{n=1}^{2} \underset{(\boldsymbol{x}_{i}^{t}, {y}_{i}^{*}) \sim\\ {(\mathcal{X}^{t},\mathcal{Y}^{*})}}{\mathbb{E}}\left[\nabla_{\theta_{fn}} \mathcal{L}^{W}_{\mathrm{ce}}\left( F_{n}\left(G\left(\boldsymbol{x}_{i}^{t}\right)\right), {y}_{i}^{*} \right)\right].
   \end{equation}
   Here $\mathcal{L}^{W}_{\mathrm{ce}}(\cdot, \cdot)$ is the weighted cross entropy loss function which is formulated as follows, 
   \begin{equation}
       \begin{aligned}
       \mathcal{L}^{W}_{ce}\left(F_{n}\left(G\left(\boldsymbol{x}_{i}^{t}\right)\right), y_{i}^{*}\right)= w_j(\boldsymbol{x}_{i}^{t}) \mathcal{L}_{ce}\left(F_{n}\left(G\left(\boldsymbol{x}_{i}^{t}\right)\right), y_{i}^{*}\right),
       \end{aligned}
   \end{equation}
   \begin{equation}
       \begin{aligned}
           w_j(\boldsymbol{x}_{i}^{t}) = 1 + e^{-E(\delta(F_n(G(\boldsymbol{x}_{i}^{t}),y_{i}^{*})))},
       \end{aligned}
   \end{equation}
   where $\delta$ represents the softmax output and $E(\cdot)$ denotes the standard information entropy. At present, we have obtained the gradient vector of source samples and target samples respectively. We minimize the discrepancy between these two gradient vectors when updating the generator in \textbf{step 3} through a gradient discrepancy loss $\mathcal{L}_{\mathrm{GD}}$, here we use cosine similarity to express the discrepancy,
   
   \begin{equation}
   \mathcal{L}_{\mathrm{GD}}=1-\frac{g_{\mathrm{s}}^{\mathrm{T}} g_{\mathrm{t}}}{\left\|g_{\mathrm{s}}\right\|_{2}\left\|g_{\mathrm{t}}\right\|_{2}},
   \end{equation}
   by which the distributions of the source domain and the target domain are aligned while the semantic information of samples is also considered.

   \subsection{Self-supervised Learning for Target Samples}  
   
   In our method, pseudo labels of target samples are indispensable to compute the gradient signal. Pseudo labeling has gained popularity in domain adaptation in recent years. Some previous studies \cite{zhang2018collaborative, choi2019pseudo, xie2018learning} directly incorporate naive pseudo labeling strategy into their methods. However, target pseudo labels that produced by the source model are still unreliable owing to the domain shift. Inspired by DeepCluster \cite{caron2018deep} and SHOT \cite{liang2020we}, in this paper, we use a weighted clustering strategy to obtain more reliable pseudo labels. Here we use softmax outputs of two classifiers to weight the samples for obtaining the centroid $c_{k}$ of the $k$-th class,
   
   \begin{equation}
   c_{k}=\frac{\sum_{n=1}^{2} \sum_{\boldsymbol{x}_{i}^{t} \in \mathcal{X}^{t}} \delta_{k}\left(F_n\left(G\left(\boldsymbol{x}_{i}^{t}\right)\right)\right) G\left(\boldsymbol{x}_{i}^{t}\right)}{\sum_{n=1}^{2} \sum_{\boldsymbol{x}_{i}^{t} \in \mathcal{X}^{t}} \delta_{k}\left(F_n\left(G\left(\boldsymbol{x}_{i}^{t}\right)\right)\right)},
   \end{equation}
   where $\delta_k$ means the corresponding $k$-th element of the softmax output $\delta$. Then pseudo labels could be obtained by the nearest centroid strategy, i.e., 
   
   \begin{equation}
   y_{i}^{*}=\arg \min _{k} d\left(G\left(\boldsymbol{x}_{i}^{t}\right), c_{k}\right),
   \end{equation}
   here $d$ could be any specific distance metric function and we use cosine distance in this paper. 
   
   In order to exploit unlabeled samples, we reinforce the \textbf{step 1} with the self-supervised learning mechanism to induce the model to learn a discriminative original target distribution and encourage each sample to lie around the correct decision boundary. The weighted classification loss $\mathcal{L}^{W}_{cls}(\cdot, \cdot)$ for self-supervised learning can be formulated as follows,
   
   \begin{equation}
       \begin{aligned}
    \mathcal{L}^{W}_{cls}\left(\mathcal{X}^{t}, \mathcal{Y}^{*}\right) = \frac{1}{2 n_{t}} \sum_{i=1}^{n_{t}} \sum_{n=1}^{2} \mathcal{L}_{ce}^W \left(F_{n}\left(G\left(\boldsymbol{x}_{i}^{t}\right)\right), y_{i}^{*}\right),
       \end{aligned}
   \end{equation}
   
   Through the above self-supervised process, we make an improvement on the discriminability of the target distribution in \textbf{step 1} by fine-tuning the model with both source and target data, so that the samples between two domains could be aligned at category-level subsequently.

   \subsection{Overall Objective and Optimization Procedure}
   
   In this work, we aim to tackle potential drawbacks of the bi-classifier adversarial learning by using self-supervised learning and minimizing the discrepancy of gradient signal produced by the source and the target domain. To sum up, by utilizing the self-supervised loss, the optimization objective of \textbf{step 1} could be reformulated as follows,
   \begin{equation}
   \begin{aligned}
       \min _{\theta_g, \theta_{f1}, \theta_{f2}} \mathcal{L}_{cls}\left(\mathcal{X}^{s}, \mathcal{Y}^{s}\right) + \alpha \mathcal{L}^{W}_{cls}\left(\mathcal{X}^{t}, \mathcal{Y}^{*}\right),
       \end{aligned}
   \end{equation}
   where $\alpha > 0$ is the trade-off parameter and could be adjusted according to the validation set. Following MCD, we use the $\ell_1$ distance to estimate the discrepancy between two distinct classifiers in \textbf{Step 2}. In \textbf{step 3}, we add a constraint on the gradient discrepancy between two domains, thus the final optimization objective becomes
   
   \begin{equation}
       \begin{aligned}
   \min _{\theta_{g}} \mathcal{L}_{dis}\left(\mathcal{X}^{t}\right) + \beta \mathcal{L}_{\mathrm{GD}}
       \end{aligned}
   \end{equation}
   where $\beta >0$ denotes the trade-off hyper-parameter. After this step, the target feature manifold will be closer to the source one while the classification accuracy of target samples is also preserved. We repeat the above process until it converges. The overall framework is shown in Fig. \ref{fig:illustration} and the training procedure is summarized in Algorithm 1.

   \section{Theoretical Analysis}
   In this section, we explain our motivation by briefly analyzing the relationship between our method and the theory of domain adaptation \cite{ben2010theory}, which gives the upper bound of the expected error on the target domain as follows, 
   \begin{equation}
      \forall h \in \mathcal{H}, R_{\mathcal{T}}(h) \leq R_{\mathcal{S}}(h)+\frac{1}{2} d_{\mathcal{H} \Delta \mathcal{H}}(\mathcal{S}, \mathcal{T})+ \lambda,
      \end{equation} 
   where $\mathcal{H}$ is the hypothesis class, $R_{\mathcal{S}}(h)$ is the expected error on the source domain which can be minimized explicitly since we have ground-truth source labels. $d_{\mathcal{H} \Delta \mathcal{H}}(\mathcal{S}, \mathcal{T})$ stands for the the domain divergence, and $\lambda$ is the error of the ideal joint hypothesis, i.e., $h^{*}=\arg \min _{h \in \mathcal{H}} R_{\mathcal{S}}(h)+R_{\mathcal{T}}(h)$. MCD employs two distinct hypotheses to reduce $d_{\mathcal{H} \Delta \mathcal{H}}(\mathcal{S}, \mathcal{T})$ and it treats $\lambda$ as a negligible constant.
   
   However, \cite{chen2020harmonizing} shows there is an optimality gap between the optimal source hypothesis and the optimal target hypothesis in UDA, if two domains are misaligned at the category-level, we can hardly find a joint hypothesis that simultaneously minimizes the source and target expected errors, leading to a large $\lambda$. Hence, our motivation is to design a model that preserves the low $d_{\mathcal{H} \Delta \mathcal{H}}(\mathcal{S}, \mathcal{T})$ in MCD and alleviates the problem caused by incorrect category-level alignment (i.e., minimizing $\lambda$). To this end, pseudo labels are necessary. Instead of using pseudo labels predicted by the source hypothesis as a direct supervision, which measures the loss of target samples \textit{w.r.t.} the trained source hypothesis rather than the expected loss of the joint hypothesis trained on them, we resort to minimizing the gradient discrepancy between domains. The insight of this trick is that we expect learning procedures of two domains not only finally yield a shared hypothesis, but also follow a similar optimization path. While this may restrict the optimization dynamics for the joint hypothesis, we argue that it enables a more fine-grained optimization and effective use of the incomplete optimizer, thus enabling a more accurate joint hypothesis and improving the accuracy of UDA.

   \begin{algorithm}[t]
    \caption{Cross-Domain Gradient Discrepancy Minimization for Unsupervised Domain Adaptation}
    \begin{algorithmic}[]
    \REQUIRE 
    The set of labeled source samples $\{\mathcal{X}^{s},\mathcal{Y}^{s}\}$ and the set of unlabeled target samples $\{\mathcal{X}^{t}\}$, the initialized generator $G$ and classifiers $F_1$ and $F_2$, the maximal epoch number $N$ and the hyper-parameter $\alpha$ and $\beta$.
    \FOR{epoch $\gets$ 1 to N} 
    \STATE \textbf{step 1:} Obtain the set of pseudo target labels $\{\mathcal{Y}^{*}\}$ through the Eq. (7) and Eq. (8). Then train $G$, $F_1$ and $F_2$ on both source and target samples:
    \begin{equation}\nonumber
        \begin{aligned}
            \min _{\theta_g, \theta_{f1}, \theta_{f2}} \mathcal{L}_{cls}\left(\mathcal{X}^{s}, \mathcal{Y}^{s}\right) + \alpha \mathcal{L}^{W}_{cls}\left(\mathcal{X}^{t}, \mathcal{Y}^{*}\right),
            \end{aligned}
    \end{equation}
    \vspace{-16pt}
    \STATE \textbf{step 2:} Train $F_1$ and $F_2$ to maximize the divergence between the outputs of two classifiers on target samples without label information, as well as preserve the accuracy on source samples:
    \begin{equation}\nonumber
        \begin{aligned}
    \min _{\theta_{f1}, \theta_{f2}} \mathcal{L}_{cls}\left(\mathcal{X}^{s}, \mathcal{Y}^{s}\right) - \mathcal{L}_{dis}\left(\mathcal{X}^{t}\right)
        \end{aligned}
    \end{equation}
    \vspace{-16pt}
    \STATE \textbf{step 3:} Train $G$ to minimize the divergence between the outputs of two classifiers with gradient similarity constraint:
    \begin{equation}\nonumber
        \begin{aligned}
    \min _{\theta_{g}} \mathcal{L}_{dis}\left(\mathcal{X}^{t}\right) + \beta \mathcal{L}_{\mathrm{GD}}
        \end{aligned}
    \end{equation}
    \vspace{-14pt}
    \ENDFOR
    \end{algorithmic}
    \end{algorithm}

   \begin{table*}[t]
    \centering
    \caption{ Accuracy(\%) on \textbf{DomainNet} dataset for unsupervised domain adaptation (ResNet-50). We evaluate all pairwise transfers among 6 domains. The column-wise fields are applied as the source domain while the row-wise fields represent the target domain.}
    \label{tab:domainnet}
    \footnotesize

    \resizebox{\textwidth}{!}{
    \begin{tabular}{c|c c c c c c c||c|c c c c c c c||c|c c c c c c c }
    \hline
    ResNet & clp & inf & pnt & qdr & rel & skt & Avg. & MCD &    clp & inf & pnt & qdr & rel & skt & Avg.  & BNM &    clp & inf & pnt & qdr & rel & skt & Avg.  \\
    \hline
    clp &     -  & 14.2& 29.6& 9.5 & 43.8& 34.3& 26.3 & clp &     -  & 15.4& 25.5& 3.3 & 44.6& 31.2& 24.0  & clp &     -  & 12.1& 33.1& 6.2 & 50.8& 40.2& 28.5  \\

    inf &    21.8&  -  & 23.2& 2.3 & 40.6& 20.8& 21.7 & inf &    24.1&  -  & 24.0& 1.6 & 35.2& 19.7& 20.9  & inf &    26.6&  -  & 28.5& 2.4 & 38.5& 18.1& 22.8\\

    pnt &    24.1& 15.0&   - & 4.6 & 45.0& 29.0& 23.5 & pnt &    31.1& 14.8&   - & 1.7 & 48.1& 22.8& 23.7  & pnt &    39.9& 12.2&   - & 3.4 & 54.5& 36.2& 29.2\\

    qdr &    12.2& 1.5 & 4.9 &  -  & 5.6 & 5.7 &  6.0 & qdr &    8.5 & 2.1 & 4.6 &  -  & 7.9 & 7.1 &  6.0  & qdr &    17.8& 1.0 & 3.6 &  -  & 9.2 & 8.3 &  8.0\\

    rel &    32.1& 17.0&36.7 &  3.6&  -  & 26.2& 23.1 & rel &    39.4& 17.8&41.2 & 1.5 &  -  & 25.2& 25.0  & rel &    48.6& 13.2&49.7 & 3.6 &  -  & 33.9& 29.8\\

    skt &    30.4& 11.3&27.8 &  3.4& 32.9&  -  & 21.2 & skt &    37.3& 12.6&27.2 &  4.1& 34.5&  -  & 23.1  & skt &    54.9& 12.8&42.3 &  5.4& 51.3&  -  & 33.3\\

    Avg. &   24.1& 11.8&24.4 &  4.7& 33.6& 23.2& 20.3 & Avg.&    28.1& 12.5&24.5 &  2.4& 34.1& 21.2& 20.5  & Avg.&    37.6& 10.3&31.4 &  4.2& 40.9& 27.3& 25.3 \\                          
    \hline
    CDAN &   clp & inf & pnt & qdr & rel & skt & Avg. & SWD &    clp & inf & pnt & qdr & rel & skt & Avg.  & CGDM&    clp & inf & pnt & qdr & rel & skt & Avg.  \\
    \hline
    clp &     -  & 13.5& 28.3& 9.3 & 43.8& 30.2& 25.0 & clp &     -  & 14.7& 31.9& 10.1& 45.3&36.5& 27.7   & clp &     -  & 16.9& 35.3& 10.8& 53.5& 36.9& 30.7  \\

    inf &    18.9&  -  & 21.4& 1.9 & 36.3& 21.3& 20.0 & inf &    22.9&  -  & 24.2& 2.5 & 33.2& 21.3& 20.0  & inf &    27.8&  -  & 28.2& 4.4 & 48.2& 22.5& 26.2\\

    pnt &    29.6& 14.4&   - & 4.1 & 45.2& 27.4& 24.2 & pnt &    33.6& 15.3&   - & 4.4 & 46.1& 30.7& 26.0  & pnt &    37.7& 14.5&   - & 4.6 & 59.4& 33.5& 30.0\\

    qdr &    11.8& 1.2 & 4.0 &  -  & 9.4 & 9.5 &  7.2 & qdr &    15.5 & 2.2 & 6.4 &  - & 11.1& 10.2&  9.1  & qdr &    14.9 & 1.5& 6.2 &  -  & 10.9& 10.2&  8.7\\

    rel &    36.4& 18.3& 40.9&  3.4&  -  & 24.6& 24.7 & rel &    41.2& 18.1 &44.2 & 4.6&  -  & 31.6& 27.9  & rel &    49.4& 20.8& 47.2& 4.8 &  -  & 38.2& 32.0\\

    skt &    38.2& 14.7&33.9 &  7.0& 36.6&  -  & 26.1 & skt &    44.2& 15.2&37.3 & 10.3&44.7 &  -  & 30.3  & skt &    50.1& 16.5& 43.7& 11.1& 55.6&  -  & 35.4\\

    Avg. &   27.0& 12.4&25.7 &  5.1& 34.3& 22.6& 21.2 & Avg.&    31.5& 13.1&28.8 &  6.4& 36.1& 26.1& 23.6  & Avg. &   36.0& 14.0& 32.1& 7.1 & 45.5& 28.3& \textbf{27.2} \\   
    \hline 
\end{tabular}}
\vspace{-13pt}
\end{table*}

   \section{Experiments}
   \subsection{Dataset Description}
   
   
   \textbf{DomainNet} \cite{peng2019moment} is the lagest and most challenging dataset to date for domain adaptation which contains about 600 thousand images distributed in 345 categories over 6 domains, including Clipart (clp), Infograph (inf), Painting (pnt), Quickdraw (qdr), Real (rel) and Sketch (skt). We adapt each domain to the other 5 domains.
   
   \textbf{VisDA-2017} \cite{peng2017visda} is a challenging large-scale dataset for UDA, which focuses on the simulation-to-reality shift. It consists of over 280K images across 12 categories. We use the training set and the validation set as the source domain and the target domain, respectively. The source domain contains 152,397 synthetic images generated by rendering 3D models and the target domain includs 55,388 real images cropped from the Microsoft COCO dataset \cite{lin2014microsoft}. 
   
   \textbf{ImageCLEF} \footnote{https://www.imageclef.org/2014/adaptation} is a standard dataset for ImageCLEF 2014 domain adaptation challenge. It is established by selecting 12 common categories shared by the following three datasets: Caltech-256 (C), ImageNet ILSVRC2012 (I) and PASCALVOC2012(P). There are 600 images in each domain and 50 for each category, whcih makes a good property for experiments.
   

   \subsection{Implementation Details}
   
   We choose PyTorch \cite{paszke2019pytorch} framework for implementing our models. NIVIDIA GeForce RTX 2080 Ti GPU is used as our hardware platform. Following MCD, we add the class balance loss in addition to the aforementioned framework to improve the diversity and accuracy in all experiments during the training procedure, the weight of which is fixed to 0.1 in this paper. For all experiments, we resize all images to $224\times224\times3$. The network architecture and the hyper-parameters setting are demonstrated as follows.    
   
   \textbf{Network Architecture.} In our experiments, we realize the generator $G$ with the ResNet-50 (for DomainNet and ImageCLEF) or ResNet-101 (for VisDA-2017) \cite{he2016deep} pre-trained on ImageNet \cite{deng2009imagenet} to extract features from raw images. Following \cite{saito2018maximum,lee2019sliced}, the original fully connected (FC) layer is replaced with a bottleneck layer of 256 units and a three-layer FC network, which are employed as our classifier $F_1$ and $F_2$, the unit number of the hidden layer is set to 1000 in all experiments. A dropout layer is utilized before each FC layer and a batch normalization (BN) layer is applied after that FC layer.
   
   \textbf{Hyper-parameters.} We train the whole network in an end-to-end fashion through back-propagation. Momentum SGD algorithm is used to optimize the network parameters with momentum 0.9 and weight decay ratio $5e^{-4}$. The classifiers are trained with learning rate $1e^{-3}$ for DomainNet and ImageCLEF, $3e^{-4}$ for VisDA-2017 since it can easily converge. The learning of the pre-trained ResNet backbone is 10 times lower that of the classifiers. The batch size is set to 32 and the trade-off parameters $\alpha$ and $\beta$ are set to 0.1 and 0.01 respectively in all experiments. Note that we do not use data augmentation such as the ten-crop ensemble used in \cite{long2018conditional,xu2019larger} during the evaluation, for fair comparison, the experimental results cited from previous works are also the versions without data augmentation.

   \begin{table*}[t]
       \centering
       \caption{Accuracy(\%) on \textbf{VisDA-2017} dataset for unsupervised domain adaptation (ResNet-101).}
       \label{tab:visda}
       \footnotesize
       \resizebox{\textwidth}{!}{
       \begin{tabular}{c|c c c c c c c c c c c c c c}
       \hline
       Method & plane & bcycl & bus & car & horse & knife & mcycl & person & plant & sktbrd & train & truck & Avg.\\
       \hline
       ResNet \cite{he2016deep} & 55.1  & 53.3 & 61.9 & 59.1& 80.6& 17.9& 79.7 & 31.2 &81.0&26.5&73.5&8.5&52.4\\
   
       DAN \cite{long2015learning}   & 87.1  & 63.0&  76.5 & 42.0 & 90.3& 42.9& 85.9&53.1&49.7&36.3&85.8&20.7&61.6 \\
   
       DANN \cite{ganin2015unsupervised}    & 81.9  & 77.7 & 82.8 & 44.3 & 81.2& 29.5& 65.1&28.6&51.9&54.6&82.8&7.8&57.4 \\
   
       MinEnt \cite{grandvalet2005semi}  & 80.3  & 75.5 & 75.8 & 48.3 & 77.9& 27.3& 69.7&40.2&46.5&46.6&79.3&16.0&57.0 \\
   
       MCD \cite{saito2018maximum}    & 87.0 & 60.9 & 83.7 & 64.0 & 88.9& 79.6& 84.7& 76.9&88.6&40.3&83.0&25.8&71.9 \\
   
       ADR \cite{saito2017adversarial}   & 87.8  & 79.5& 83.7 & 65.3 & 92.3& 61.8& 88.9&73.2&87.8&60.0&85.5&32.3&74.8 \\
   
       SWD \cite{lee2019sliced}  &  90.8  & 82.5& 81.7 & 70.5 & 91.7& 69.5& 86.3& 77.5&87.4&63.6&85.6&29.2&76.4  \\
   
       CDAN+E \cite{long2018conditional}  & 85.2  & 66.9&83.0 & 50.8 & 84.2& 74.9& 88.1&74.5&83.4&76.0&81.9&38.0&73.9 \\
   
       AFN \cite{xu2019larger}   & \textbf{93.6}  & 61.3& \textbf{84.1} & \textbf{70.6} & \textbf{94.1}& 79.0& \textbf{91.8}& 79.6&89.9&55.6&89.0&24.4&76.1 \\
    
       BNM \cite{cui2020towards}   & 89.6  & 61.5& 76.9 & 55.0 & 89.3& 69.1& 81.3& 65.5&90.0&47.3&\textbf{89.1}&30.1&70.4 \\
       
       MSTN+DSBN \cite{chang2019domain} & 94.7 & 86.7 & 76.0 & 72.0 & 95.2 & 75.1 & 87.9 & 81.3 & 91.1 & 68.9 & 88.3 & 45.5 & 80.2 \\

       \hline
       CGDM (ours)  & 93.4  & \textbf{82.7}& 73.2 & 68.4 & 92.9 & \textbf{94.5} & 88.7& \textbf{82.1}&\textbf{93.4}&\textbf{82.5}&86.8&\textbf{49.2}&\textbf{82.3} \\
       \hline 

   \end{tabular}}
   \vspace{-17pt}
   \end{table*}
   
    \vspace{-10pt}
       \begin{figure*}[t]
           \centering
           \label{fig:Visualization}
           \subfigure[ResNet-50]{
           \includegraphics[width=0.24\linewidth] {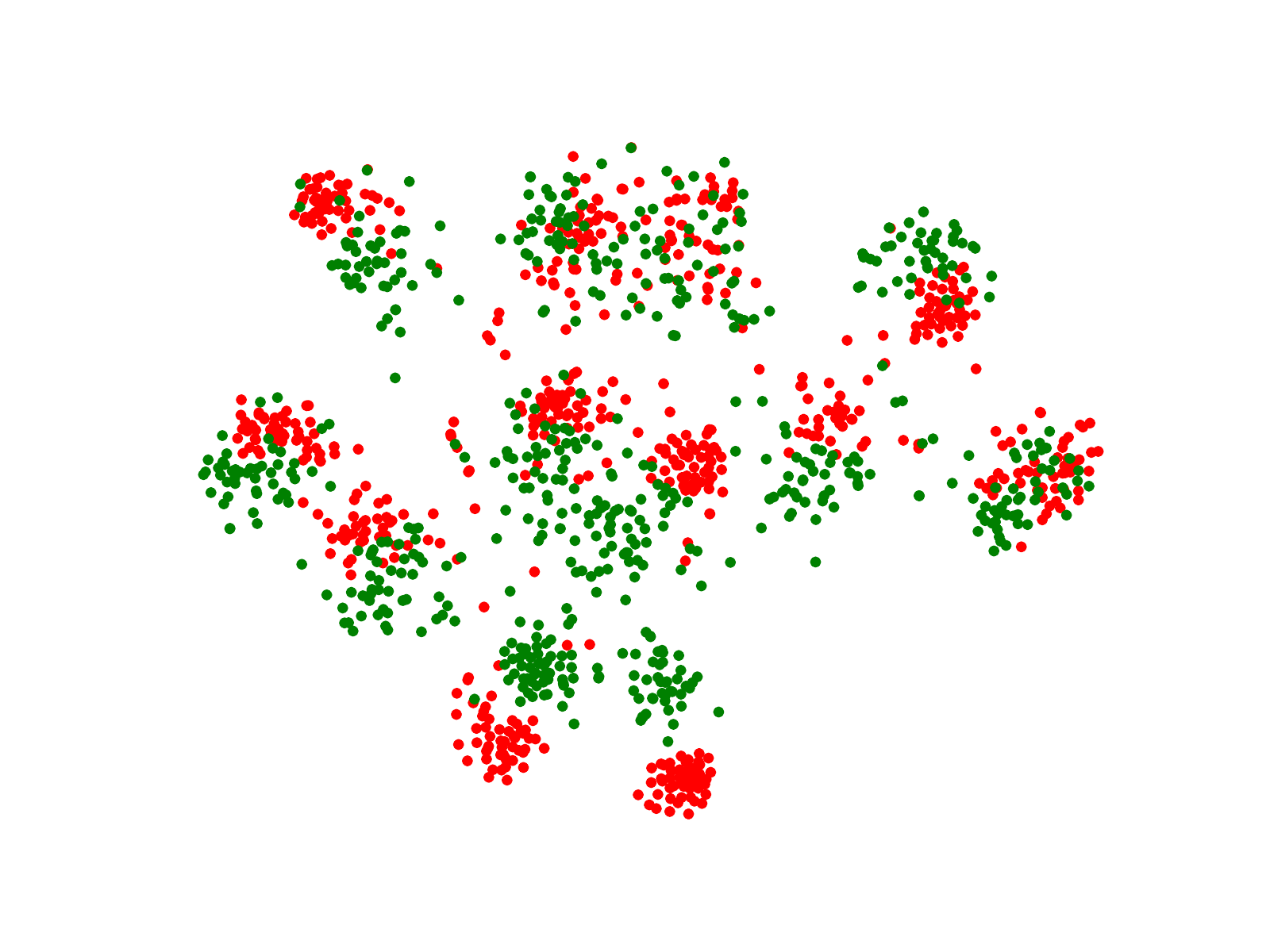}
           \label{fig:Visualization_src}
           }
            \subfigure[DANN]{
               \includegraphics[width=0.24\linewidth] {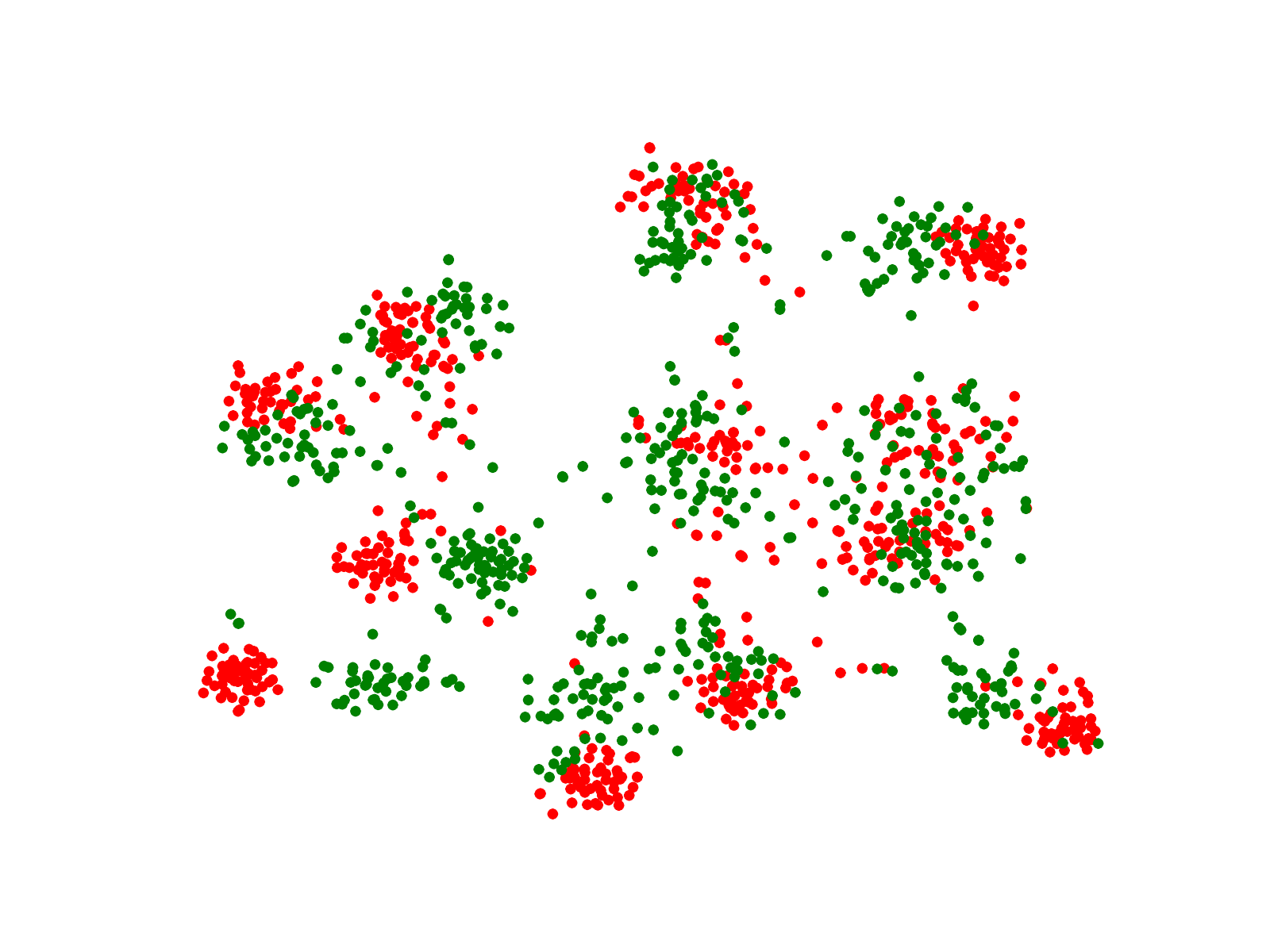}
               \label{fig:Visualization_DANN}
               }
            \subfigure[MCD]{
            \includegraphics[width=0.23\linewidth] {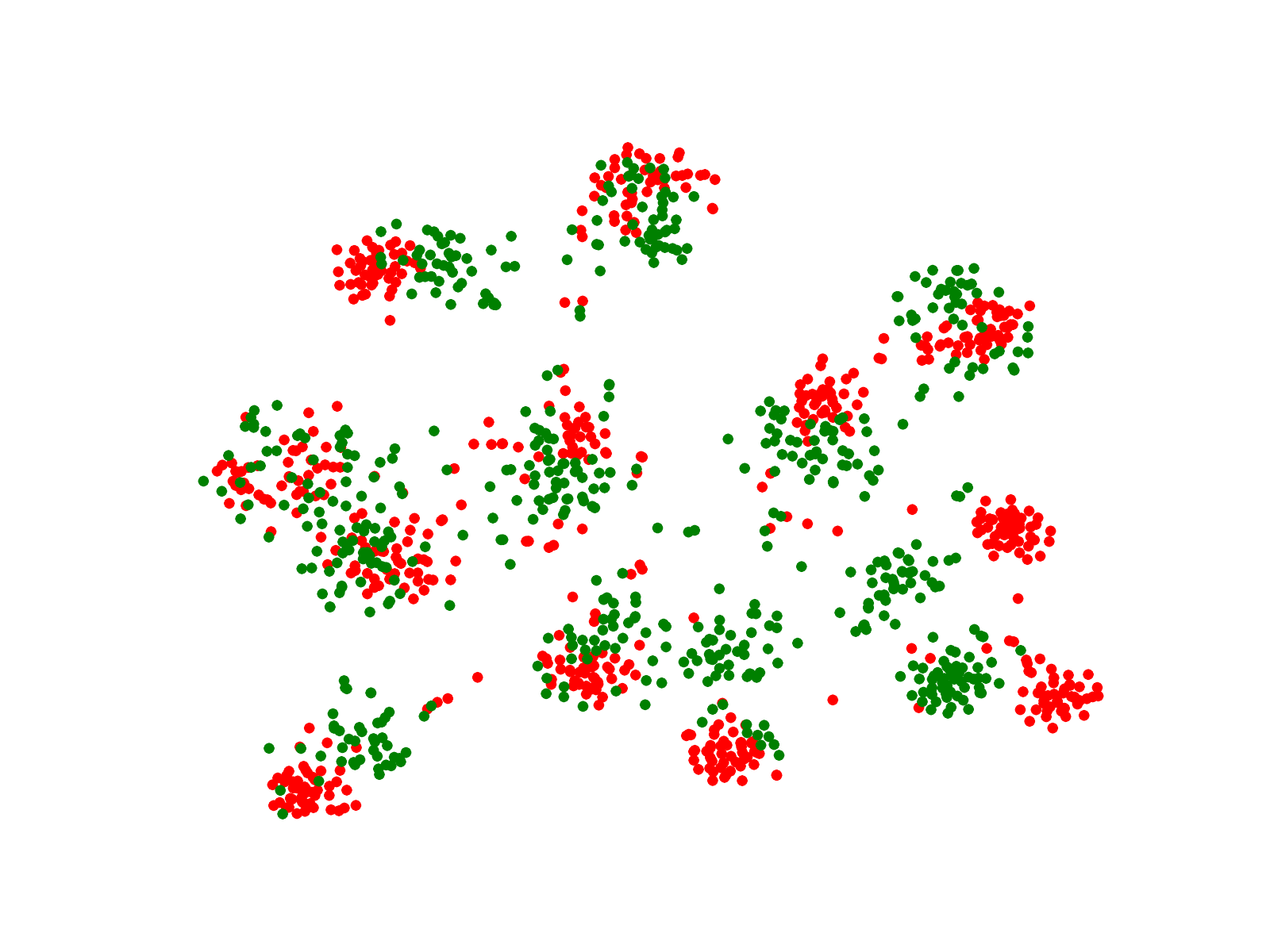}
            \label{fig:Visualization_mcd}
            } 
           \subfigure[Ours]{
           \includegraphics[width=0.23\linewidth] {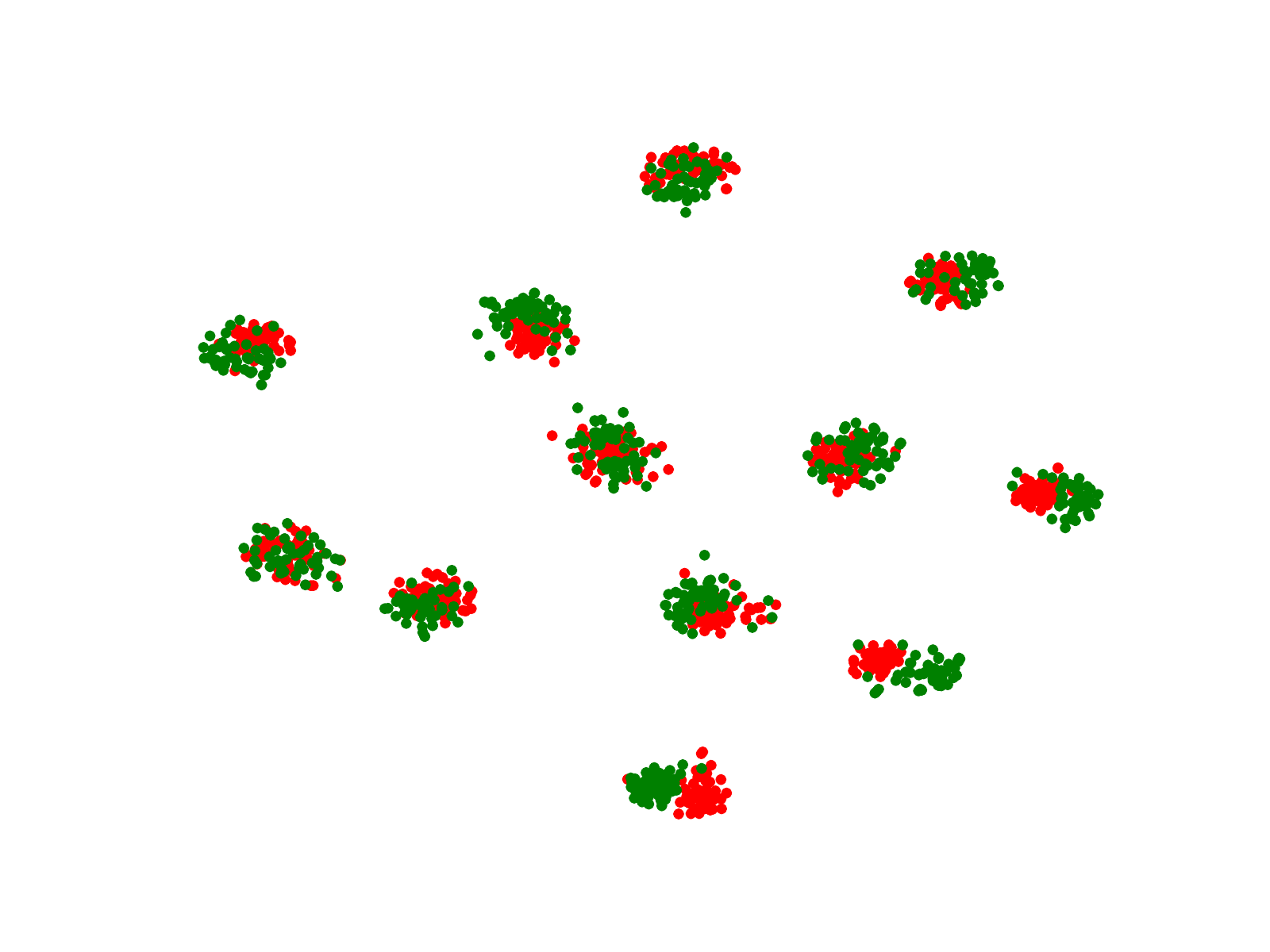}
           \label{fig:Visualization_ours}
           }
           \caption{Visualization of features using t-SNE. We take C (red) $\rightarrow$ I (green) on ImageCLEF as an example.}
           \vspace{-15pt}
       \end{figure*}

       \begin{figure}[t]
           \centering
           \subfigure[Classification Loss]{
           \includegraphics[width=0.46\linewidth] {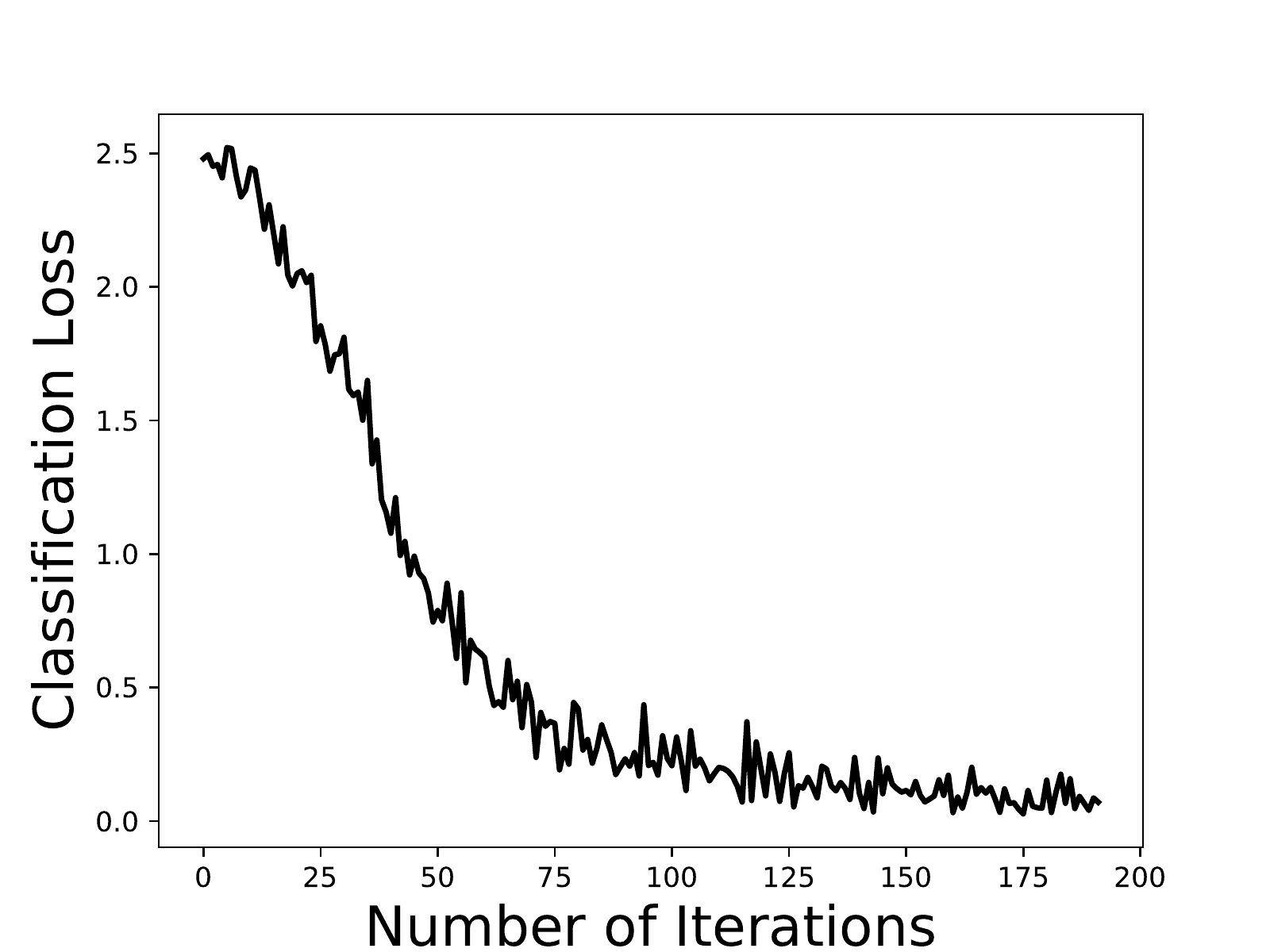}
           \label{fig:convergence}
           }
           \subfigure[Target Accuracy]{
           \includegraphics[width=0.46\linewidth] {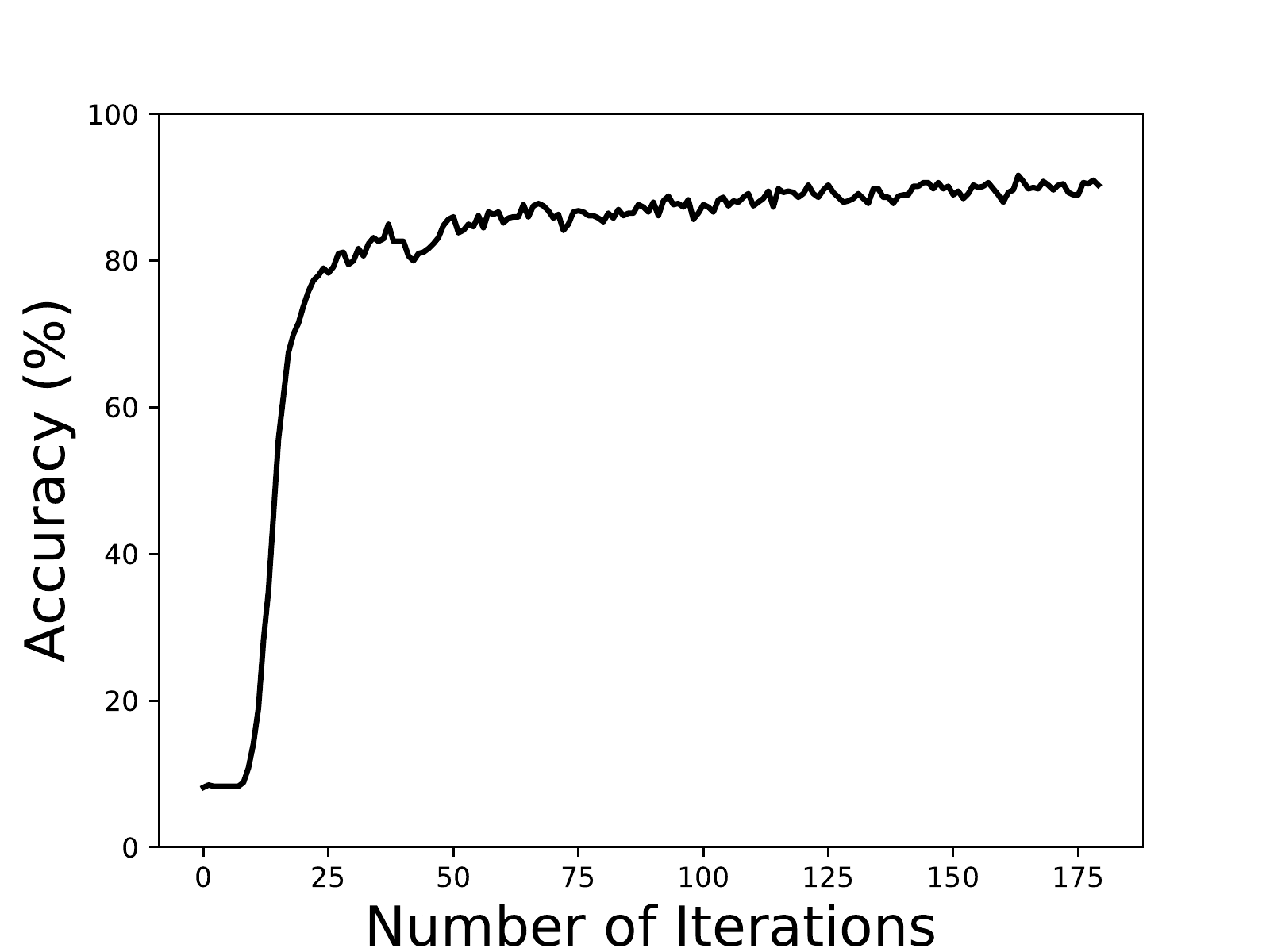}
           \label{fig:accuracy}
           }
           \caption{Training process. We take C $\rightarrow$ I on ImageCLEF as an example. The left figure depicts the classification loss during training. The right figure reports the accuracy of the target domain during the training.}
           \vspace{-15pt}
           \end{figure}
       
           \begin{figure}[t]
               \centering
               \subfigure[Parameter Sensitivity]{
               \includegraphics[width=0.46\linewidth] {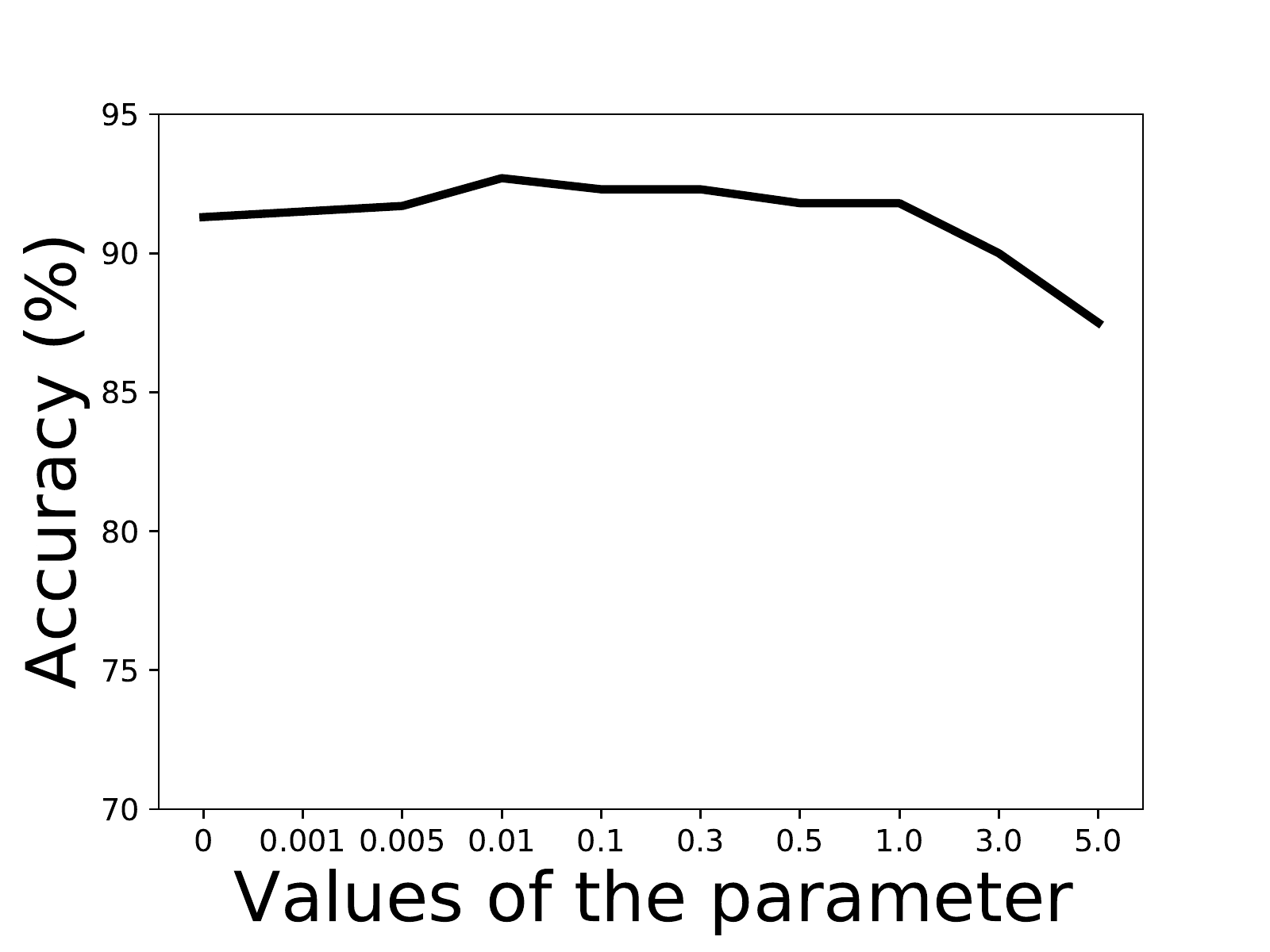}
               \label{fig:parameter}
               }
               \subfigure[Ablation Analysis]{
               \includegraphics[width=0.46\linewidth] {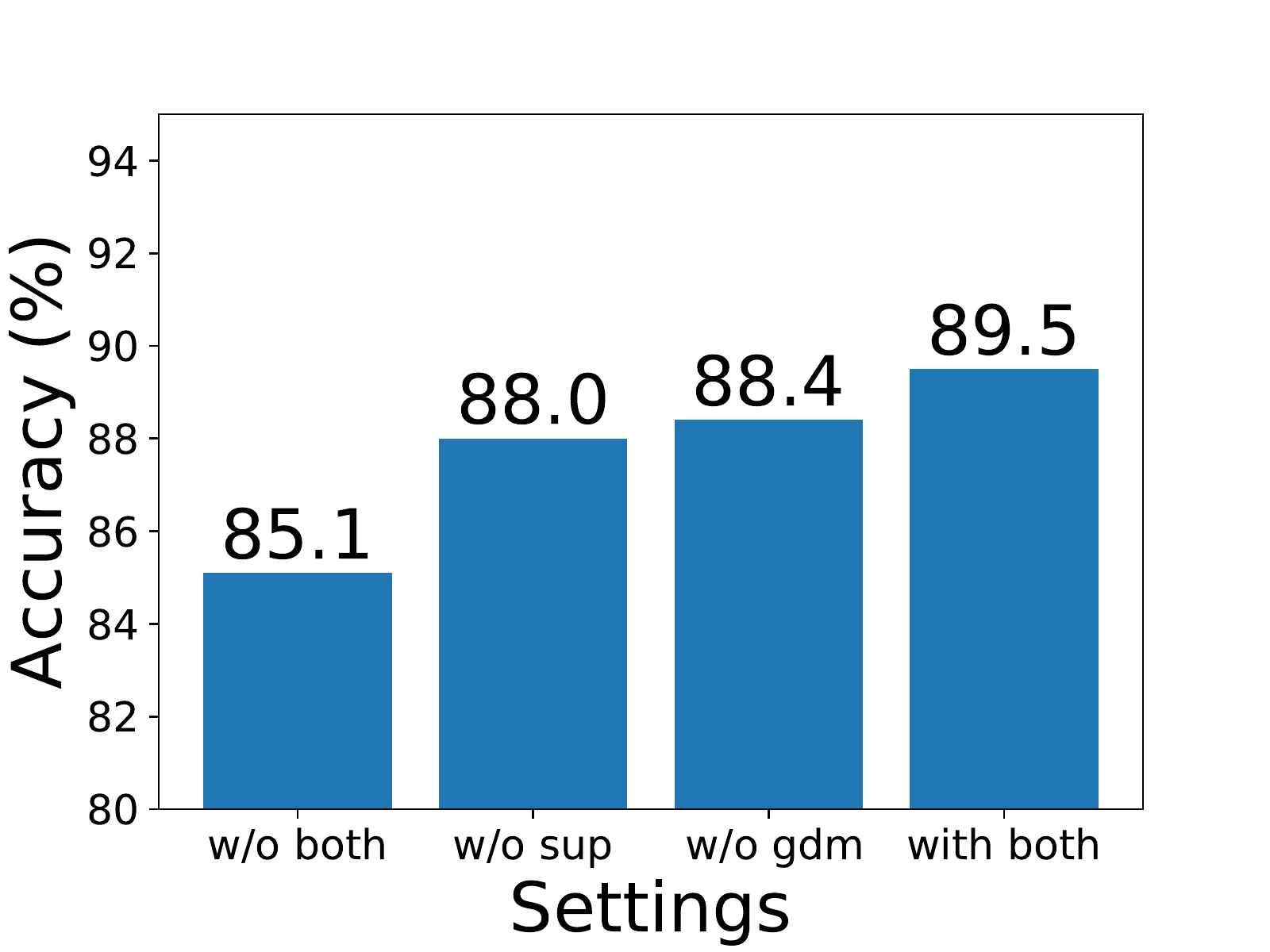}
               \label{fig:ablation}
               }
               \label{myfigure}
               \caption{Model analysis. The left figure shows the parameter sensitivity of our model (We take the task C $\rightarrow$ I as an example). While the right figure shows the ablation study on ImageCLEF. The w/o is short for without, sup for self-supervised learning and gdm for gradient discrepancy minimization, respectively.}
               \vspace{-10pt}
           \end{figure}

           \begin{table*}[t]
            \centering
            \caption{Accuracy(\%) on \textbf{ImageCLEF} dataset for unsupervised domain adaptation (ResNet-50).}
            \label{tab:imageclef}
            \small
            \setlength{\tabcolsep}{3.8mm}{
            \begin{tabular}{c c c c c c c c}
            \hline
            Method (Source $\rightarrow$ Target) & I $\rightarrow$ P & P $\rightarrow$ I & I $\rightarrow$ C & C $\rightarrow$ I & C $\rightarrow$ P & P $\rightarrow$ C & Avg \\
            \hline
            ResNet \cite{he2016deep} & 74.8$\pm$0.3  & 83.9$\pm$0.1 & 91.5$\pm$0.3 & 78.0$\pm$0.2 & 65.5$\pm$0.3& 91.2$\pm$0.3& 80.7 \\
        
            DAN \cite{long2015learning}& 74.5$\pm$0.4  &  82.2$\pm$0.2& 92.8$\pm$0.2 & 86.3$\pm$0.4 & 69.2$\pm$0.4& 89.80$\pm$0.4& 82.5 \\
        
            RTN \cite{long2016unsupervised}   & 75.6  & 86.8 & 95.3 & 86.9 & 72.7& 92.2& 84.9 \\
        
            DANN \cite{ganin2015unsupervised}  & 75.0$\pm$0.6  & 86.0$\pm$0.3 & 96.2$\pm$0.4 & 87.0$\pm$0.5 & 74.3$\pm$0.5& 91.5$\pm$0.6& 85.0 \\
        
            MinEnt \cite{grandvalet2005semi} & 76.2  & 85.7& 93.5 & 83.5 & 69.3& 89.7& 83.0 \\
        
            MCD \cite{saito2018maximum}   & 77.3  & 89.2& 92.7 & 88.2 & 71.0& 92.3& 85.1 \\
        
            CDAN+E \cite{long2018conditional}  & 77.7$\pm$0.3  & 90.7$\pm$0.2& \textbf{97.7}$\pm$0.3 & 91.3$\pm$0.3 & 74.2$\pm$0.2& 94.3$\pm$0.3& 87.7 \\
        
            AFN \cite{xu2019larger}    & \textbf{79.3}$\pm$0.1  & \textbf{93.3}$\pm$0.4&96.3$\pm$0.4 & 91.7$\pm$0.0 & 77.6$\pm$0.1& 95.3$\pm$0.1& 88.9 \\
        
            BNM \cite{cui2020towards}   & 77.2  & 91.2& 96.2 & 91.7 & 75.7& \textbf{96.7}& 88.1 \\
            \hline
            CGDM (ours)  & 78.7$\pm$0.2  &\textbf{93.3}$\pm$0.1& 97.5$\pm$0.3 & \textbf{92.7}$\pm$0.2  & \textbf{79.2}$\pm$0.1 & 95.7$\pm$0.2 & \textbf{89.5} \\
            \hline 
        \end{tabular}}
        \vspace{-12pt}
        \end{table*}
   \vspace{5pt}    
   \subsection{Results}
   Here we show the comparison between our CGDM framework and other well-known UDA baselines, especially the methods that are most related to our work (e.g. MCD and SWD), to verify that our formulation can significantly boost the accuracy \textit{w.r.t.} to these baselines. Note that in our experiments, the classification results are obtained by averaging the outputs of two distinct classifiers. The quantitative results are described as follows. 
   
   \textbf{Results on DomainNet.} In this experiment, we transfer each domain to the other five domains, and the class-wise average accuracy of each adaptation is recported in Table \ref{tab:domainnet}. From Table \ref{tab:domainnet}, our CGDM evidently supresses other mainstream domain adaptation methods in mean accuracy. 
   
   The results are obtained after 10 epochs. In particular, both MCD and SWD perform significantly poorer than our method with 6.7\% and 3.6\% lower mean accuracy. The low accuracy of the source only model can infer that original target samples are likely to be distributed in a mess facing such a challenging dataset with a large number of categories. Although the target samples outside the suport of the source can be detected by the classifiers, they seem like to be matched to wrong decision boundaries, resulting in the wrong category-level alignment. 
   
   For other adversarial methods, they just forcibly align distributions and neglect the inherent gap between the source domain and the target domain, which may result in negative transfer. Our method applies a clustering-based self-supervised mechanism to improve the original accuracy of target samples, and the discrepancy between gradients of two domains is minimized to induce the generator to learn towards the direction of accurate classification for both source domain and target domain samples. The results on Table \ref{tab:domainnet} suggest that our model can still achieve a more accurate adaptation in the target domain when facing such a large-scale dataset. 
   
   \textbf{Results on VisDA-2017.} We obtain the results on VisDA-2017 dataset after 10 epochs. The results in Table \ref{tab:visda} illustrate our method can still outperform other popular approaches even if there is a large domain gap between synthetic and real images. Specifically, our method performs much better than the source only model in all categories with the improvement up to 29.9\% in terms of mean accuracy. For MCD and SWD which also apply the bi-classifier adversarial learning paradigm, our method also significantly supresses them with the improvement of 10.4\% and 5.9\% respectively. In general, our method achieves the best in 5 categories: knife, person, plant, sktbrd and truck. In more difficult categories such as knife, sktbrd and truck, our method is much better than other existing methods, achieving 94.5\%, 82.5\% and 49.2\% respectively. These excellent results strongly demonstrate the advantage of our method in improving the accuracy of the target domain.

   \textbf{Results on ImageCLEF.} We access 6 types of adaptation scenarios on ImageCLEF and Table \ref{tab:imageclef} reports the experimental results. We stop the training process after 100 epochs. The experimental results of the comparative methods are cited from previous papers, sine some of them do not reported the randomness on this dataset, so we do not report their randomness too. Table \ref{tab:imageclef} shows that our method outperforms other popular baselines and achieves the best average accuracy (89.5\%). As for the difficult scenarios (e.g. C $\rightarrow$ P), our method can obviously improve the accuracy of the target domain, which shows the effectiveness of our approach. 
   
   In addition, in scenarios that involve many categories (DomainNet) or there is a large domain gap (VisDA-2017), our model outperforms other methods by a large margin. We can empirically conclude that the the consideration of accuracy is particularly important in these scenarios.

   \subsection{Model Analysis}
   
   In this section, we take a further step to analyze the properties of the model in terms of convergence, feature visualization, parameter sensitivity and ablation study.
   
   \textbf{Feature Visualization.} To further demonstrate the effectiveness of our model and have an intuitive understanding, we visualize the feature distribution learned by the model using t-SNE \cite{maaten2008visualizing}. We take C $\rightarrow$ I on ImageCLEF as an example. It shows that in all scenarios, the source samples present a discriminative distribution. While in the absence of domain adaptation, the target samples likely to be disorganized, as shown in Fig. \ref{fig:Visualization_src}. Fig. \ref{fig:Visualization_DANN} shows that using the domain adversarial neural network, the distribution discrepancy between the source domain and the target domain is reduced, while the discriminability of the target domain is relatively poor. Fig. \ref{fig:Visualization_mcd} shows that MCD improves the discriminability of features. After using our method, the distribution of target samples aligns well with the source one at category-level. The result verifies the effectiveness and feasibility of our method.

   \textbf{Convergence.} In order to verify the convergence tendency of our method, we report the classification loss on source samples in Fig. \ref{fig:convergence}, this loss is obtained on the classifier $F_1$. Fig. \ref{fig:accuracy} depicts the accuracy curve of target samples during the training process. Here we take C $\rightarrow$ I on ImageCLEF as an example. These figures show that our method can markedly reduce the loss and improve the accuracy with the number of iterations increases, which proves that the training process is smooth and convergent.
   
   \textbf{Sensitivity to Hyper-Parameter.} We check the hyper-parameter sensitivity of our model. Our framework involves two hyper-parameters $\alpha$ and $\beta$, the gradient discrepancy loss constitutes the main new methodological contribution in this paper while the self-supervised loss is used as an auxiliary loss, so we fix the $\alpha$ as 0.1 because we empirically find that it performs well in this value. Then we report the accuracy of C $\rightarrow$ I by turning $\beta$ form 0 to 5, which is shown in Fig. \ref{fig:parameter}. When $\beta$ is between 0 and 1, there was no obvious deterioration in accuracy, when $\beta$ is larger than 1, the accuracy begins to decline since the weight of the gradient discrepancy loss is larger than that of the supervised learning loss on source samples. In general, our model is not sensitive to parameters. 
   
   \textbf{Ablation Study.} Our framework is composed of a self-supervised learning module and a gradient discrepancy minimization module. In order to test the effectiveness of each module, we conduct experiments without the self-supervised learning and the gradient discrepancy minimization, respectively. Limited by space, we take the mean accuracy on ImageCLEF as an example. As shown in Fig. \ref{fig:ablation}, both two modules can significantly improve the accuracy compared to the vanilla bi-classifier adversarial learning, and the model performs better when they work together.
   
   \textbf{Conditional Gradient Discrepancy.} We use two types of gradient discrepancy in this paper. In addition to align marginal gradients by regarding all samples in a batch as a whole, we have tried to align gradients for source and target samples within the same category separately. However, in our experiments, we could not see obvious improvement using the latter version. It indicates that the overall gradient of a batch is able to express the category information of the domain, thus we use the former one for less computation.
   
   \section{Conclusion}
   
   In this paper, we investigate the inaccurate issue of the conventional bi-classifier adversarial learning for domain adaptation. To alleviate this issue, we propose a novel UDA method which aims to minimize the gradient discrepancy between two domains to achieve a better distribution alignment at category-level. In addition, self-supervised learning is used to obtain more reliable pseudo labels of target samples. Extensive experimental results on large scale datasets demonstrate the advantage of our method. 

   \section*{Acknowledgement}
   This work was supported in part by the National Natural Science Foundation of China under Grant 61806039, 62073059 and 61832001, and in part by Sichuan Science and Technology Program under Grant 2020YFG0080. 

{\small
\bibliographystyle{ieee_fullname}
\bibliography{refer}

\begin{thebibliography}{10}\itemsep=-1pt

\bibitem{ben2010theory}
Shai Ben-David, John Blitzer, Koby Crammer, Alex Kulesza, Fernando Pereira, and
  Jennifer~Wortman Vaughan.
\newblock A theory of learning from different domains.
\newblock {\em Machine learning}, 79(1):151--175, 2010.

\bibitem{caron2018deep}
Mathilde Caron, Piotr Bojanowski, Armand Joulin, and Matthijs Douze.
\newblock Deep clustering for unsupervised learning of visual features.
\newblock In {\em Proceedings of the European Conference on Computer Vision
  (ECCV)}, pages 132--149, 2018.

\bibitem{chang2019domain}
Woong-Gi Chang, Tackgeun You, Seonguk Seo, Suha Kwak, and Bohyung Han.
\newblock Domain-specific batch normalization for unsupervised domain
  adaptation.
\newblock In {\em Proceedings of the IEEE/CVF Conference on Computer Vision and
  Pattern Recognition}, pages 7354--7362, 2019.

\bibitem{chen2020harmonizing}
Chaoqi Chen, Zebiao Zheng, Xinghao Ding, Yue Huang, and Qi Dou.
\newblock Harmonizing transferability and discriminability for adapting object
  detectors.
\newblock In {\em Proceedings of the IEEE/CVF Conference on Computer Vision and
  Pattern Recognition}, pages 8869--8878, 2020.

\bibitem{chen2019transferability}
Xinyang Chen, Sinan Wang, Mingsheng Long, and Jianmin Wang.
\newblock Transferability vs. discriminability: Batch spectral penalization for
  adversarial domain adaptation.
\newblock In {\em International Conference on Machine Learning}, pages
  1081--1090, 2019.

\bibitem{choi2019pseudo}
Jaehoon Choi, Minki Jeong, Taekyung Kim, and Changick Kim.
\newblock Pseudo-labeling curriculum for unsupervised domain adaptation.
\newblock {\em arXiv preprint arXiv:1908.00262}, 2019.

\bibitem{cui2020towards}
Shuhao Cui, Shuhui Wang, Junbao Zhuo, Liang Li, Qingming Huang, and Qi Tian.
\newblock Towards discriminability and diversity: Batch nuclear-norm
  maximization under label insufficient situations.
\newblock In {\em Proceedings of the IEEE/CVF Conference on Computer Vision and
  Pattern Recognition}, pages 3941--3950, 2020.

\bibitem{deng2009imagenet}
Jia Deng, Wei Dong, Richard Socher, Li-Jia Li, Kai Li, and Li Fei-Fei.
\newblock Imagenet: A large-scale hierarchical image database.
\newblock In {\em 2009 IEEE conference on computer vision and pattern
  recognition}, pages 248--255. Ieee, 2009.

\bibitem{ganin2015unsupervised}
Yaroslav Ganin and Victor Lempitsky.
\newblock Unsupervised domain adaptation by backpropagation.
\newblock In {\em International conference on machine learning}, pages
  1180--1189, 2015.

\bibitem{ganin2016domain}
Yaroslav Ganin, Evgeniya Ustinova, Hana Ajakan, Pascal Germain, Hugo
  Larochelle, Fran{\c{c}}ois Laviolette, Mario Marchand, and Victor Lempitsky.
\newblock Domain-adversarial training of neural networks.
\newblock {\em The Journal of Machine Learning Research}, 17(1):2096--2030,
  2016.

\bibitem{grandvalet2005semi}
Yves Grandvalet and Yoshua Bengio.
\newblock Semi-supervised learning by entropy minimization.
\newblock In {\em Advances in neural information processing systems}, pages
  529--536, 2005.

\bibitem{he2016deep}
Kaiming He, Xiangyu Zhang, Shaoqing Ren, and Jian Sun.
\newblock Deep residual learning for image recognition.
\newblock In {\em Proceedings of the IEEE conference on computer vision and
  pattern recognition}, pages 770--778, 2016.

\bibitem{kang2019contrastive}
Guoliang Kang, Lu Jiang, Yi Yang, and Alexander~G Hauptmann.
\newblock Contrastive adaptation network for unsupervised domain adaptation.
\newblock In {\em Proceedings of the IEEE/CVF Conference on Computer Vision and
  Pattern Recognition}, pages 4893--4902, 2019.

\bibitem{lee2019sliced}
Chen-Yu Lee, Tanmay Batra, Mohammad~Haris Baig, and Daniel Ulbricht.
\newblock Sliced wasserstein discrepancy for unsupervised domain adaptation.
\newblock In {\em Proceedings of the IEEE Conference on Computer Vision and
  Pattern Recognition}, pages 10285--10295, 2019.

\bibitem{li2019cycle}
Jingjing Li, Erpeng Chen, Zhengming Ding, Lei Zhu, Ke Lu, and Zi Huang.
\newblock Cycle-consistent conditional adversarial transfer networks.
\newblock In {\em Proceedings of the 27th ACM International Conference on
  Multimedia}, pages 747--755, 2019.

\bibitem{li2020maximum}
Jingjing Li, Erpeng Chen, Zhengming Ding, Lei Zhu, Ke Lu, and Heng~Tao Shen.
\newblock Maximum density divergence for domain adaptation.
\newblock {\em IEEE transactions on pattern analysis and machine intelligence},
  2020.

\bibitem{li2019locality}
Jingjing Li, Mengmeng Jing, Ke Lu, Lei Zhu, and Heng~Tao Shen.
\newblock Locality preserving joint transfer for domain adaptation.
\newblock {\em IEEE Transactions on Image Processing}, 28(12):6103--6115, 2019.

\bibitem{li2021faster}
Jingjing Li, Mengmeng Jing, Hongzu Su, Ke Lu, Lei Zhu, and Heng~Tao Shen.
\newblock Faster domain adaptation networks.
\newblock {\em IEEE Transactions on Knowledge and Data Engineering}, 2021.

\bibitem{li2018heterogeneous}
Jingjing Li, Ke Lu, Zi Huang, Lei Zhu, and Heng~Tao Shen.
\newblock Heterogeneous domain adaptation through progressive alignment.
\newblock {\em IEEE transactions on neural networks and learning systems},
  30(5):1381--1391, 2018.

\bibitem{li2018transfer}
Jingjing Li, Ke Lu, Zi Huang, Lei Zhu, and Heng~Tao Shen.
\newblock Transfer independently together: A generalized framework for domain
  adaptation.
\newblock {\em IEEE transactions on cybernetics}, 49(6):2144--2155, 2018.

\bibitem{li2020deep}
Shuang Li, Chi~Harold Liu, Qiuxia Lin, Qi Wen, Limin Su, Gao Huang, and
  Zhengming Ding.
\newblock Deep residual correction network for partial domain adaptation.
\newblock {\em IEEE Transactions on Pattern Analysis and Machine Intelligence},
  2020.

\bibitem{liang2020we}
Jian Liang, Dapeng Hu, and Jiashi Feng.
\newblock Do we really need to access the source data? source hypothesis
  transfer for unsupervised domain adaptation.
\newblock {\em arXiv preprint arXiv:2002.08546}, 2020.

\bibitem{lin2014microsoft}
Tsung-Yi Lin, Michael Maire, Serge Belongie, James Hays, Pietro Perona, Deva
  Ramanan, Piotr Doll{\'a}r, and C~Lawrence Zitnick.
\newblock Microsoft coco: Common objects in context.
\newblock In {\em European conference on computer vision}, pages 740--755.
  Springer, 2014.

\bibitem{long2015learning}
Mingsheng Long, Yue Cao, Jianmin Wang, and Michael Jordan.
\newblock Learning transferable features with deep adaptation networks.
\newblock In {\em International conference on machine learning}, pages 97--105.
  PMLR, 2015.

\bibitem{long2018conditional}
Mingsheng Long, Zhangjie Cao, Jianmin Wang, and Michael~I Jordan.
\newblock Conditional adversarial domain adaptation.
\newblock In {\em Advances in Neural Information Processing Systems}, pages
  1640--1650, 2018.

\bibitem{long2016unsupervised}
Mingsheng Long, Han Zhu, Jianmin Wang, and Michael~I Jordan.
\newblock Unsupervised domain adaptation with residual transfer networks.
\newblock In {\em Advances in neural information processing systems}, pages
  136--144, 2016.

\bibitem{long2017deep}
Mingsheng Long, Han Zhu, Jianmin Wang, and Michael~I Jordan.
\newblock Deep transfer learning with joint adaptation networks.
\newblock In {\em International conference on machine learning}, pages
  2208--2217, 2017.

\bibitem{luo2019taking}
Yawei Luo, Liang Zheng, Tao Guan, Junqing Yu, and Yi Yang.
\newblock Taking a closer look at domain shift: Category-level adversaries for
  semantics consistent domain adaptation.
\newblock In {\em Proceedings of the IEEE/CVF Conference on Computer Vision and
  Pattern Recognition}, pages 2507--2516, 2019.

\bibitem{maaten2008visualizing}
Laurens van~der Maaten and Geoffrey Hinton.
\newblock Visualizing data using t-sne.
\newblock {\em Journal of machine learning research}, 9(Nov):2579--2605, 2008.

\bibitem{pan2010domain}
Sinno~Jialin Pan, Ivor~W Tsang, James~T Kwok, and Qiang Yang.
\newblock Domain adaptation via transfer component analysis.
\newblock {\em IEEE Transactions on Neural Networks}, 22(2):199--210, 2010.

\bibitem{paszke2019pytorch}
Adam Paszke, Sam Gross, Francisco Massa, Adam Lerer, James Bradbury, Gregory
  Chanan, Trevor Killeen, Zeming Lin, Natalia Gimelshein, Luca Antiga, et~al.
\newblock Pytorch: An imperative style, high-performance deep learning library.
\newblock In {\em Advances in neural information processing systems}, pages
  8026--8037, 2019.

\bibitem{pei2018multi}
Zhongyi Pei, Zhangjie Cao, Mingsheng Long, and Jianmin Wang.
\newblock Multi-adversarial domain adaptation.
\newblock {\em arXiv preprint arXiv:1809.02176}, 2018.

\bibitem{peng2019moment}
Xingchao Peng, Qinxun Bai, Xide Xia, Zijun Huang, Kate Saenko, and Bo Wang.
\newblock Moment matching for multi-source domain adaptation.
\newblock In {\em Proceedings of the IEEE International Conference on Computer
  Vision}, pages 1406--1415, 2019.

\bibitem{peng2017visda}
Xingchao Peng, Ben Usman, Neela Kaushik, Judy Hoffman, Dequan Wang, and Kate
  Saenko.
\newblock Visda: The visual domain adaptation challenge.
\newblock {\em arXiv preprint arXiv:1710.06924}, 2017.

\bibitem{saito2017adversarial}
Kuniaki Saito, Yoshitaka Ushiku, Tatsuya Harada, and Kate Saenko.
\newblock Adversarial dropout regularization.
\newblock {\em arXiv preprint arXiv:1711.01575}, 2017.

\bibitem{saito2018maximum}
Kuniaki Saito, Kohei Watanabe, Yoshitaka Ushiku, and Tatsuya Harada.
\newblock Maximum classifier discrepancy for unsupervised domain adaptation.
\newblock In {\em Proceedings of the IEEE Conference on Computer Vision and
  Pattern Recognition}, pages 3723--3732, 2018.

\bibitem{tzeng2017adversarial}
Eric Tzeng, Judy Hoffman, Kate Saenko, and Trevor Darrell.
\newblock Adversarial discriminative domain adaptation.
\newblock In {\em Proceedings of the IEEE conference on computer vision and
  pattern recognition}, pages 7167--7176, 2017.

\bibitem{wang2017instance}
Rui Wang, Masao Utiyama, Lemao Liu, Kehai Chen, and Eiichiro Sumita.
\newblock Instance weighting for neural machine translation domain adaptation.
\newblock In {\em Proceedings of the 2017 Conference on Empirical Methods in
  Natural Language Processing}, pages 1482--1488, 2017.

\bibitem{xie2018learning}
Shaoan Xie, Zibin Zheng, Liang Chen, and Chuan Chen.
\newblock Learning semantic representations for unsupervised domain adaptation.
\newblock In {\em International conference on machine learning}, pages
  5423--5432. PMLR, 2018.

\bibitem{xu2019larger}
Ruijia Xu, Guanbin Li, Jihan Yang, and Liang Lin.
\newblock Larger norm more transferable: An adaptive feature norm approach for
  unsupervised domain adaptation.
\newblock In {\em Proceedings of the IEEE International Conference on Computer
  Vision}, pages 1426--1435, 2019.

\bibitem{yan2019weighted}
Hongliang Yan, Zhetao Li, Qilong Wang, Peihua Li, Yong Xu, and Wangmeng Zuo.
\newblock Weighted and class-specific maximum mean discrepancy for unsupervised
  domain adaptation.
\newblock {\em IEEE Transactions on Multimedia}, 2019.

\bibitem{zellinger2017central}
Werner Zellinger, Thomas Grubinger, Edwin Lughofer, Thomas Natschl{\"a}ger, and
  Susanne Saminger-Platz.
\newblock Central moment discrepancy (cmd) for domain-invariant representation
  learning.
\newblock {\em arXiv preprint arXiv:1702.08811}, 2017.

\bibitem{zhang2018collaborative}
Weichen Zhang, Wanli Ouyang, Wen Li, and Dong Xu.
\newblock Collaborative and adversarial network for unsupervised domain
  adaptation.
\newblock In {\em Proceedings of the IEEE Conference on Computer Vision and
  Pattern Recognition}, pages 3801--3809, 2018.

\bibitem{zhang2019domain}
Yabin Zhang, Hui Tang, Kui Jia, and Mingkui Tan.
\newblock Domain-symmetric networks for adversarial domain adaptation.
\newblock In {\em Proceedings of the IEEE/CVF Conference on Computer Vision and
  Pattern Recognition}, pages 5031--5040, 2019.

\bibitem{zou2018unsupervised}
Yang Zou, Zhiding Yu, BVK Kumar, and Jinsong Wang.
\newblock Unsupervised domain adaptation for semantic segmentation via
  class-balanced self-training.
\newblock In {\em Proceedings of the European conference on computer vision
  (ECCV)}, pages 289--305, 2018.

\end{thebibliography}
}

\end{document}